\documentclass[runningheads,a4paper]{llncs}
\pdfoutput=1
    
    \usepackage{amssymb}
    \usepackage{graphicx}
    \usepackage{url}
    \usepackage[vlined]{algorithm2e}
    \usepackage{amsmath}
    \usepackage{csquotes}
    \usepackage{listings}
    \usepackage[absolute,overlay]{textpos}

    \newcommand{\keywords}[1]{\par\addvspace\baselineskip
    \noindent\keywordname\enspace\ignorespaces#1}
    
\begin{document}
    
    \mainmatter  
    
    \title{Modelling, simulation, and planning for the MoleMOD system}
    \titlerunning{}
    
    \author{Michaela Brejchov\'a\inst{1} \and Miroslav Kulich\inst{2}\orcidID{0000-0002-0997-5889} \and Jan Petr\v{s}\inst{3} \and Libor P\v{r}eu\v{c}il\inst{2}%
    \authorrunning{Michaela Brejchov\'a \and Miroslav Kulich \and Jan Petr\v{s} \and Libor P\v{r}eu\v{c}il}
    }
    
    
    \institute{
    Faculty of Electrical Engineering\\
    Czech Technical University in Prague, Prague, Czech Republic\\
    brejcmi3@fel.cvut.cz\and
        Czech Institute of Informatics, Robotics, and Cybernetics,\\
        Czech Technical University in Prague, Prague, Czech Republic\\
        kulich@cvut.cz, preucil@cvut.cz\\
        \url{http://imr.ciir.cvut.cz} \and
    Faculty of Architecture\\
    Czech Technical University in Prague, Prague, Czech Republic\\
    petrjan@fa.cvut.cz\\
    }
    
    %
    %
    
    \maketitle
    
\begin{textblock*}{\textwidth}(1in+\hoffset+\oddsidemargin,1cm) 
\centering
\small
In: Modelling and Simulation for Autonomous Systems (MESAS 2018). Cham: Springer International Publishing AG, 2019. p. 3-15. ISSN 0302-9743. ISBN 978-3-030-14983-3.
\end{textblock*}

    \begin{abstract}

        MoleMOD is a heterogeneous self-reconfigurable modular robotic system to be employed in architecture and civil engineering.  
        In this paper we present two components of the MoleMOD infrastructure - a test environment and a planning algorithm. The test environment for simulation and visualization of active parts as well as passive blocks of MoleMOD is based on Gazebo - a powerful general-purpose robotic simulator. The key effort has been put into preparation of realistic models of passive and active components taking into account their physical characteristics.  Moreover, given a starting configuration of the MoleMOD system and a final configuration an approach to plan collision-free trajectories for a fleet of active parts is introduced.

   \end{abstract}

\section{Introduction}
MoleMOD is a unique self-reconfigurable modular robotic system developed at Czech Technical University in Prague~\cite{Petrs2017,molemod}. The system is heterogeneous as it consists of active robots and passive modular building blocks and it is inspired by colonies like termites, ants or bees, which permanently rebuild and adapt their ``houses'' to surroundings and current conditions.

The robots are flexible, can rotate, and are able to connect to the passive blocks as well as pick up and carry them along a given trajectory. This approach offers extensive possibilities of reconfiguration and adaption due to separated mobile and passive units. The two parts separation gave the system its acronym Mole (animal) + MOD (module). The passive part can be imagined as units of regular 3D lattice, just like individual crystals or voxels, respectively, in a crystalline lattice or virtual digitized volume. 
        
The active part, as it is now, is further decomposed to three essential parts: soft/flexible body, revolving heads and a rotator in the centre. The head is primarily used for screwing the passive units to hold together and the  secondary function is to ride over the construction site1. The flexible body allows for the peristaltic movement, through the trajectories within the passive block conglomerate. A secondary function of the body, not least important, is the manipulation with the blocks, so picking up and carrying. Finally, the rotator allows rotation of the blocks as is typical for majority of modular robotic systems.

\begin{figure}[tbp]
	\centering
	\includegraphics[width=0.8\textwidth]{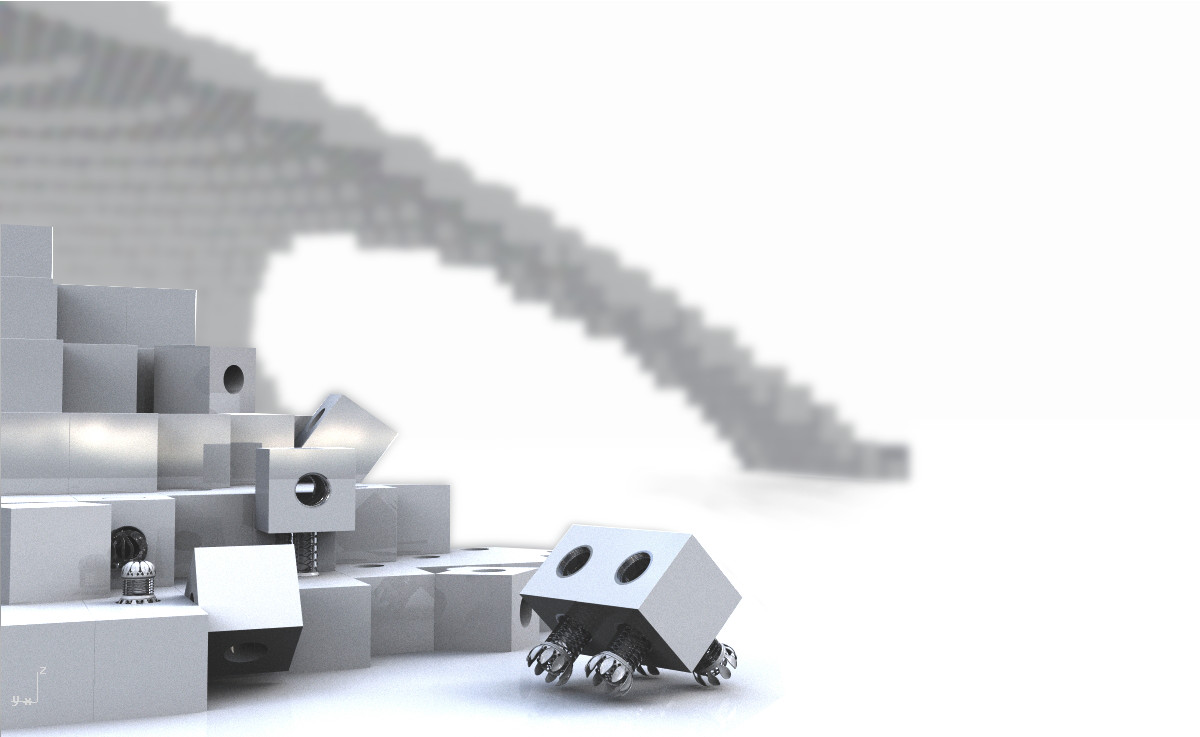}
	\caption{The MoleMOD system.}
\end{figure}
        
The MoleMOD system is very adaptable and can be used in many various situations, especially at locations that are not safe for people, or there is a problem to build -- places like deserts, mountains or polar regions, which cannot be inhabited, but it can be necessary to build there. MoleMOD may not be used only for building houses, but also for bridges, pylons or research stations.

The system does not need cranes or other external construction machines. Therefore it is quite easy to transport it to the building site. It will even be possible to transfer only robots; building modules will be created by a 3D printer from local materials. This way the system could be used for the colonization of another planet.

Other uses may be for example temporary constructions. Tribunes for sports events, such as the Olympic Games or the World Cup races, markets, exhibitions, festivals, events that last only a few days or weeks. No less important is the possibility of using the system in case of a disaster. It can be building of bridges after a flood, shelters for people who have lost their homes due to a catastrophe and so on. Also, it can be used after a nuclear disaster, when the presence of humans is not possible because of radiation.        

\begin{figure}[htb]
	\centering
	\includegraphics[width=0.8\textwidth]{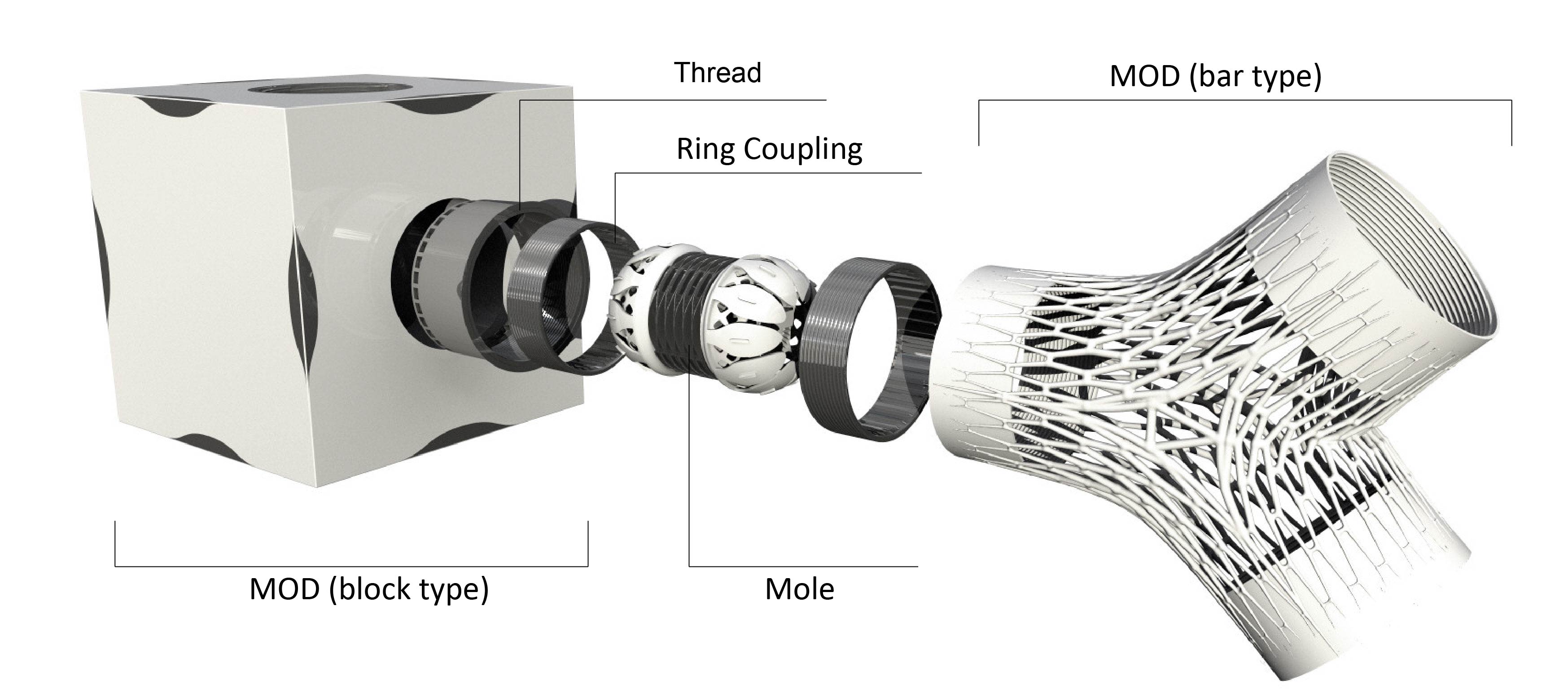}
	\caption{Structure of the system.}
\end{figure}

The rest of the paper is organized as follows. 
Section~\ref{sec:gazebo} is dedicated to the description of the simulation environment for MoleMOD based on the Gazebo simulator, while the planning approach is presented in Section~\ref{sec:planning}.
The final comments and future work are described in Section~\ref{sec:conclusion}.

\section{Simulation environment}
\label{sec:gazebo}

This section provides a description of the Gazebo simulator and design of models (both passive blocks and active parts) for testing the MoleMOD system in Gazebo. 
Gazebo is an open-source robotics simulator, which can be used to design robots, test algorithms and artificial intelligence systems using realistic scenarios. 
The simulator offers indoor and outdoor environments with the possibility of setting several properties, such as wind, gravitation, friction and so on. Gazebo includes multiple physics engines (ODE,Bullet, Simbody and DART), a library of robot models and environments, several types of sensors and functional graphical and programmatic interfaces~\cite{g}.

A simulation environment in Gazebo is described in so called world files, which include specification of elements such as robots, lights, sensors or static objects. The files use SDF (Simulation Description Format), an XML format originally developed for Gazebo.

Model files are similar to world files but contain only specifications for a model. The model created by this file can be included in a world file, so it is possible to use one model several times without rewriting the entire code. Also, there is the online model database.

SDF models can be just simple shapes but also complex robots. Basically, a model consists of links, joints, sensors, collision objects, visuals and plugins~\cite{SDF}.
A~\textbf{link} contains the physical properties. It is a body of the model or its part. It may have many collision and visual elements.
A~\textbf{collision} element is a geometry that is used to check collisions. A link can contain many collisions.
A~\textbf{visual} element visualize parts of a link. A link can have many visuals or none.
A~\textbf{joint} connects two links. Each joint has a parent and a child, an axis of rotation and some other properties.
A~\textbf{sensor} collects data from the world and these are then used by plugins.

\subsection{Modelling passive blocks}
\begin{figure}[htb]
	\centering
	\includegraphics[height=0.44\textwidth]{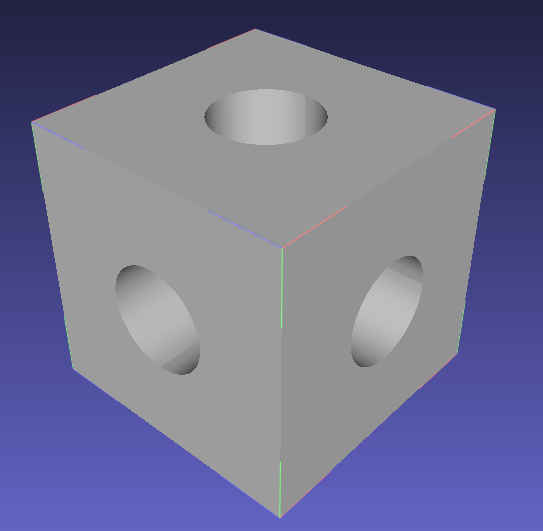}
    \includegraphics[height=0.44\textwidth]{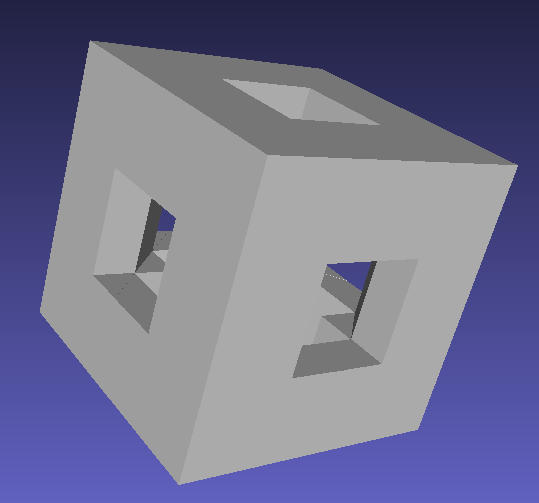}
	\caption{A cube-shaped block with two tunnels. Precise approximation of a cube-shaped block with circular tunnels contains 6174 triangular faces (left), while a description of a block with squared tunnels contains 352 faces.}
    \label{fig:cube}
\end{figure}

As passive blocks do not move, only their shape needs to be defined in the form of a triangular mesh --  a collection of triangles. 
Even simplest blocks, cubes with two straight circular tunnels, consist of many triangles, when a precise approximation is needed, see Fig.~\ref{fig:cube}~(left). We, therefore, assume square-shaped tunnels for which only a fraction of triangles is needed, Fig.\ref{fig:cube}~(right).  

Triangular meshes can be defined in Collada (Collaborative Design Activity) files (.dae), which can be created either manually or using some modelling tool like MeshLab (\url{www.meshlab.net/}) or Blender (\url{www.blender.org}). This can be imported into a world file similarly to Listing~\ref{lst:collada}. 

\begin{lstlisting}[basicstyle=\footnotesize\ttfamily,caption={A world file with mesh importing},label={lst:collada}, captionpos=b,numbers=none,numberstyle=\tiny,frame=lines,tabsize=2]
    <sdf version="1.4">
        <world name="default">
            <include>
                <uri>model://sun</uri>
            </include>
            <include>
                <uri>model://ground_plane</uri>
            </include>
            <model name="my_mesh">
                <pose>0 0 .2 0 0 0</pose>
                <link name='tunnel'>
                    <visual name='visual'>
                        <transparency>0.5</transparency>
                        <geometry>
                            <mesh>
                                <uri>file://1.dae</uri>
                                <scale>1 1 1</scale>
                            </mesh>
                        </geometry>
                    </visual>
                    <collision name='collision'>
                        <geometry>
                            <mesh>
                                <uri>file://1.dae</uri>
                                <scale>1 1 1</scale>
                            </mesh>
                        </geometry>
                    </collision>
                </link>		
            </model>
        </world>
    </sdf>
    \end{lstlisting}
    
    \subsection{Modelling active parts}
    
    Active parts -- robots -- consist of three components: a soft body, revolving heads and a rotator. 
   Precise moddeling of these will be unnecessarily complex which will significantly slow down simulation of the whole MoleMOD system. Gazebo has, moreover, no ways how to simulate soft components.  
   We thus model the robots making use of components available in Gazebo -- links and joints. 
   To describe the modelling process, we start with a simple model with limited functionality and make the model more complex and powerfull subsequently in several steps.
    
    \subsubsection{The basic model}
    
\begin{figure}[htbp]
        \centering
        \includegraphics[height=0.2\textwidth]{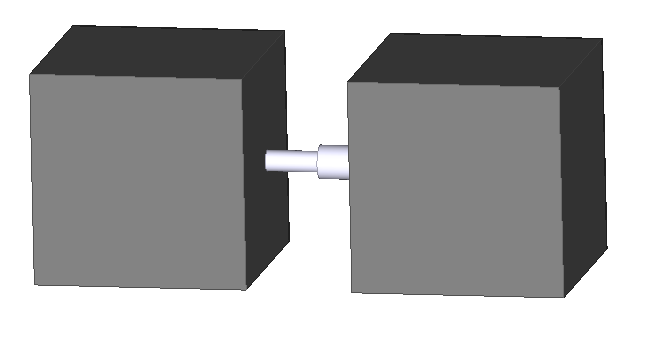}
        \caption{The basic model}
        \label{fig:model1}
    \end{figure}

    The basic model is constructed from two cubes connected by a prismatic joint, see Fig.~\ref{fig:model1}. The model is limited to move only in one direction forwards or backwards by expanding and contracting the joint and changing the frictions of the links.
    
    The forward motion of the model consists of four parts:
    \begin{enumerate}
        \item setting frictions
        \item the joint expansion
        \item setting frictions
        \item the joint contraction
    \end{enumerate}
        
    The frictions of the links are set in a model plugin. Before moving itself, it is necessary to lower the friction of the front cube (the first cube in the direction of the movement). For expanding the joint, a positive velocity is set to the joint in the plugin which leads to  movement of the front cube.

    The simulator updates in fixed intervals. When a predefined number of iterations is done, the plugin changes the sign of the velocity and swaps frictions – higher for the front cube and lower for the back cube. After repeating the same number of iterations, the joint position is set to the initial (contracted) state.
    
    \subsubsection{The rotating model}
    
\begin{figure}[htbp]
        \centering
        \includegraphics[height=0.2\textwidth]{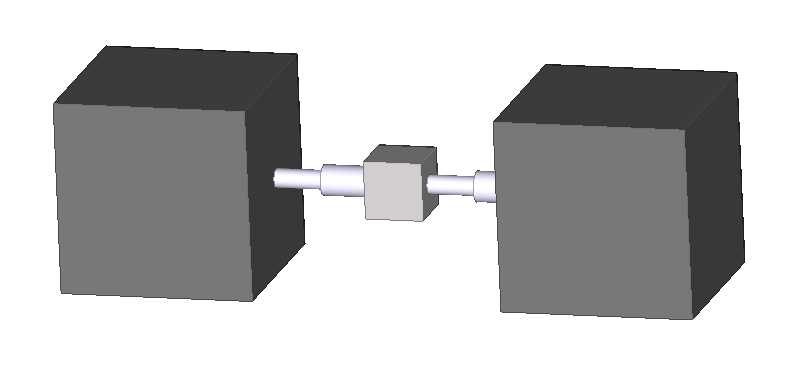}
        \caption{The rotating model}
        \label{fig:model2}
    \end{figure}

    For turning, it is needed to add a revolute joint in the middle of the model. Because the simulator cannot connect two joints, it is necessary to add two links and one more prismatic joint. Now the model is made of two head cubes, two small middle cubes, two prismatic joints and one revolute joint, Fig.~\ref{fig:model2}.
    
    The first prismatic joint connects the first head cube with the first middle cube. This cube is joined with the second middle cube by the revolute joint. These both cubes are immaterial and they are placed on themselves. The second middle cube is connected to the second cube with the second prismatic joint.
    
    After these changes, the code of the plugin has to be adapted. For translation, both prismatic joints expand and contract at the same time.
    
    Rotation is similar to translation. The four parts of motion are:
    \begin{enumerate}
        \item setting frictions
        \item the joint rotation to the direction, where we want to turn
        \item setting frictions
        \item the joint rotation to the different direction
    \end{enumerate}
    
    At first, the friction of the cube to be shifted is reduced. The plugin then sets an input velocity to the revolute joint. This velocity can be positive or negative depending on the direction of the turn.
    Finally, frictions are swapped and velocity with an opposite sign is set to the joint. In contrast to the moving ahead, this motion stops after one step and will never be repeated, because the model would spin on the same place.
    
    \subsubsection{Translation/rotation controller}
    Unfortunately, the movements of the previous models are not precise enough. The joints move for the same time, but it does not guarantee that their final position is the same. To make the motion more accurate, a simple controller has been designed. The input to the controller is a position of the joint -– the length of the prismatic joint or the angle of the rotation of the revolute joint.
    
    \begin{lstlisting}[basicstyle=\ttfamily,caption={A joint controller},captionpos=b,numbers=left,numberstyle=\tiny,frame=lines,tabsize=4,label={lst:control}]
    if(position < required_value - accuracy) {
        setVel(vel);
    } else if(position > required_value + accuracy) {
        setVel(-vel);
    } else {
        setVel(0.0);
    }
    \end{lstlisting}
    
    Listing\ref{lst:control} shows a primitive controller that sets a positive velocity to the joint, if its position is lower than the required one, negative if larger or zero if it is in the interval determined by the deviation.
    
    For better control, we can divide the possible positions into more intervals. The result will be similar to the example above; it will only contain more conditions.   
    The controller used in the simulator is divided into five intervals. At the beginning of the motion, the joint is set to an initial speed. When the joint position is close to the required position, the speed is decreased to half the value.
    
    The controller accuracy is 50 micrometres. In this range, the joint is set to zero velocity. If the position is larger or smaller (depending on the direction of the motion) the plugin sets a negative speed to the joint. Thanks to this condition there is no need to set the reverse speed for the contracting, the controller will solve it. Also, it is not necessary to count updates; one step consists of expanding to input length and contracting back to the initial state. 
    
    \begin{figure}[htbp]
        \centering
        \includegraphics[width=\textwidth]{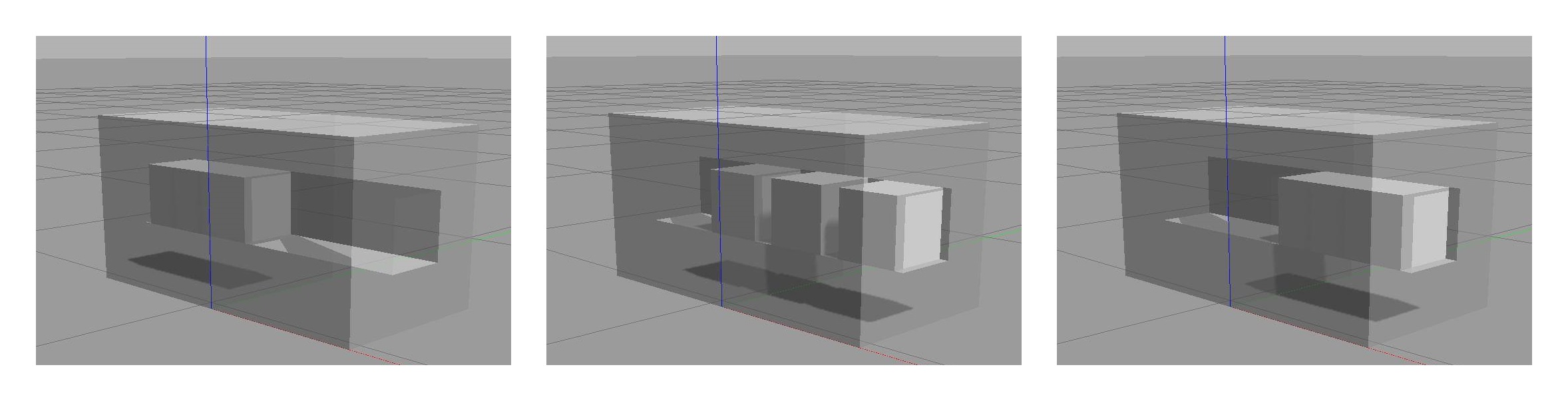}
        \caption{Moving in a tunnel - the final model}
    \end{figure}
    
    The second input for ahead motion is a number of iterations, but in this case, it is not the number of updates, but the number of calls of the controller (the total sum of all expanding and contracting). The distance the model moves is equal to the product of one step length and the number of iterations.
    
    \subsubsection{The final model}
    
    The model with just one revolute joint is not sufficient. To turn in a tunnel or lift a building unit, at least two revolute joints are needed. It is also not convenient to use a simple revolute joint, because it can rotate only around one axis. With the joints, the robot could rotate sideways or up and down, but could not do both of these operations. A solution to this problem is to use another joint. The joint is called universal in the simulator and it can rotate around two axes.
    
\begin{figure}[htbp]
        \centering
        \includegraphics[height=0.2\textwidth]{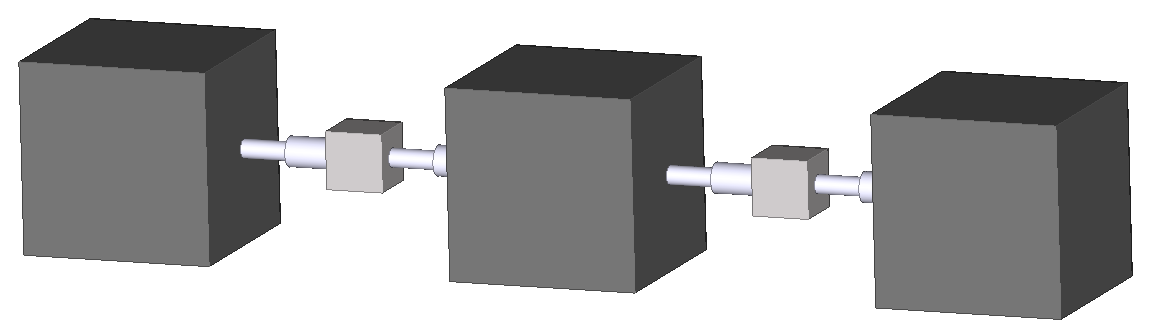}
        \caption{The final model}
        \label{fig:final_model}
    \end{figure}

    The final model consists of three main cubes, four prismatic joints, two universal joints and four little immaterial cubes that are between joints as depicted in Fig.~\ref{fig:final_model}.
    
\subsection{Experiments}

The models for the simulator have been made; the next step is to test them. In the previous chapter, the model of the MoleMOD robot was introduced and two simple moves (forward moving and rotation) were described.

\begin{figure}[htbp]
	\centering
	\includegraphics[width=\textwidth]{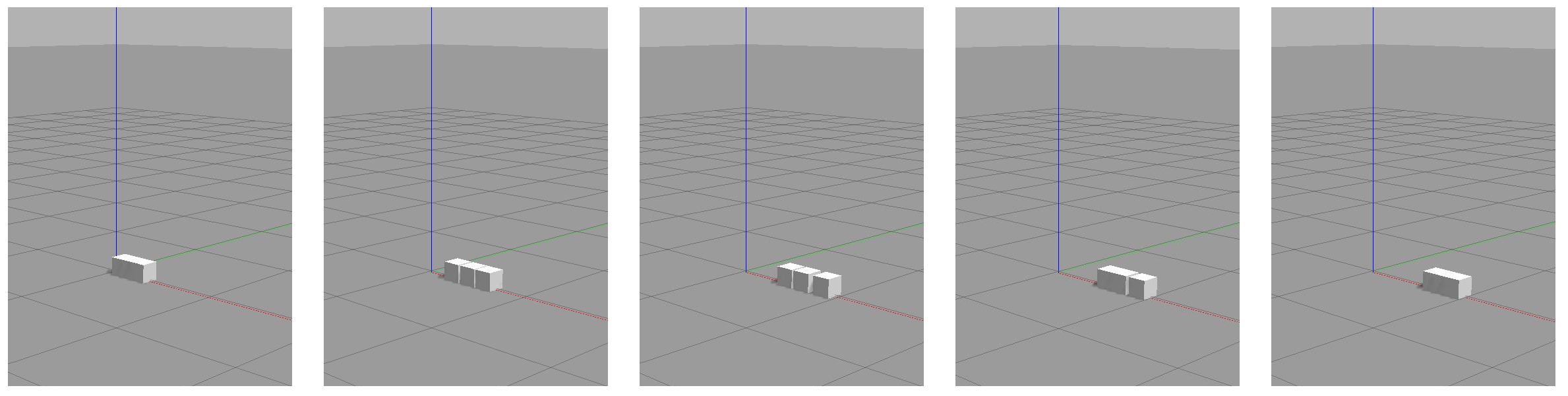}
    \caption{Forward moving}
    \label{fig:forward}
\end{figure}

Forward movement works on the same principle that has been described previously, see Fig.~\ref{fig:forward}. The only difference is that the final model has four prismatic joints instead of two. The model could use all four joints when moving, but it is simpler to use only two, for example, the first and the fourth joint. That will spare us larger changes in the code and also the motion will be more accurate.

\begin{figure}[htbp]
	\centering
	\includegraphics[width=\textwidth]{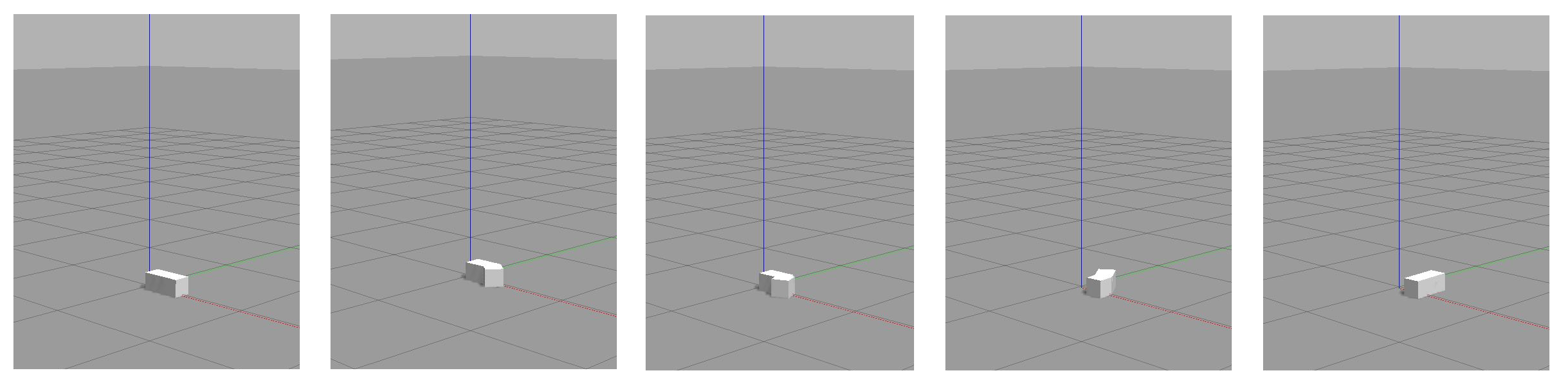}
    \caption{Turning right}
    \label{fig:turn_right}
\end{figure}

The rotation remains practically the same with only two minor changes. The last model contains two universal joints, so it is needed to decide which joint will rotate. Then the rotation axis has to be set because the universal joint can rotate around two axes. An example of rotation is depicted in Fig.~\ref{fig:turn_right}.

However, for simulating the work of the system, these two motions are not sufficient. For the basic version of the planning, it is essential to add lifting (putting a module on a next block or lifting a module just up), shifting modules and moves of robots in tunnels.

\begin{figure}[htbp]
	\centering
	\includegraphics[width=\textwidth]{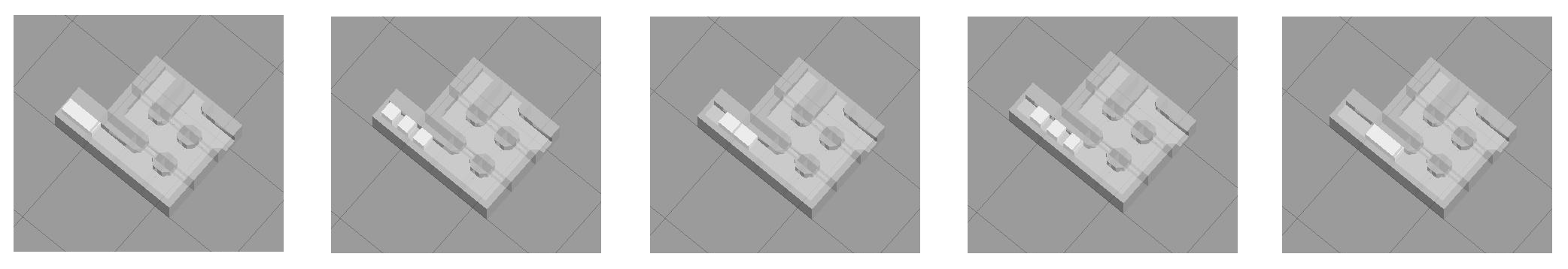}
    \caption{Moving in the straight tunnel}
    \label{fig:tunel}
\end{figure}

Robot movement in a straight tunnel is exactly the same as the forward moving. The length of the motion is adjusted according to the size of the block, see Fig.~\ref{fig:tunel}. 

\begin{figure}[htbp]
	\centering
	\includegraphics[width=\textwidth]{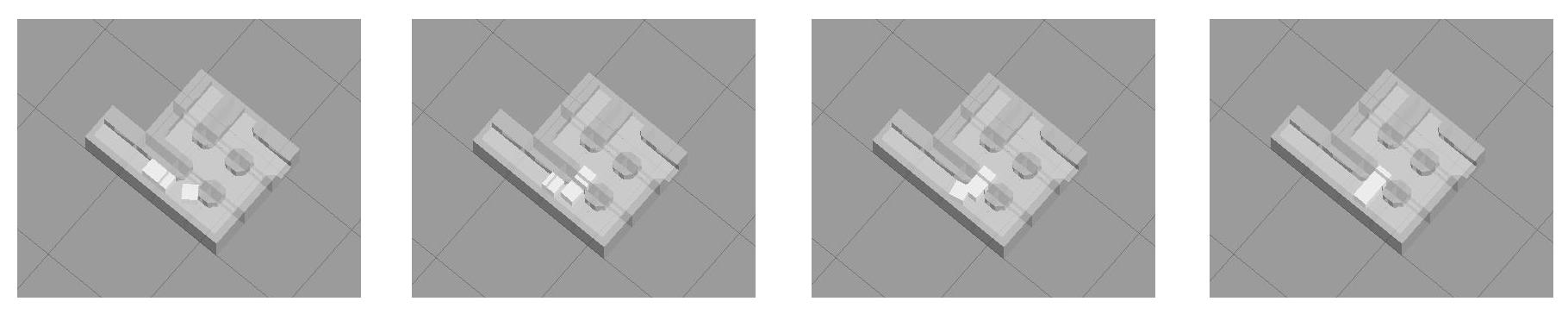}
    \caption{Turning in the module}
    \label{fig:turn_tunel}
\end{figure}

Turning in a tunnel is more complicated. Because the robot is only a little bit smaller than the tunnel, it is impossible to turn around at once. It is needed to combine both types of moves - rotation and translation. This can be achieved by using a joint controller, see an example in Fig.~\ref{fig:turn_tunel}

\begin{figure}[htbp]
	\centering
	\includegraphics[width=\textwidth]{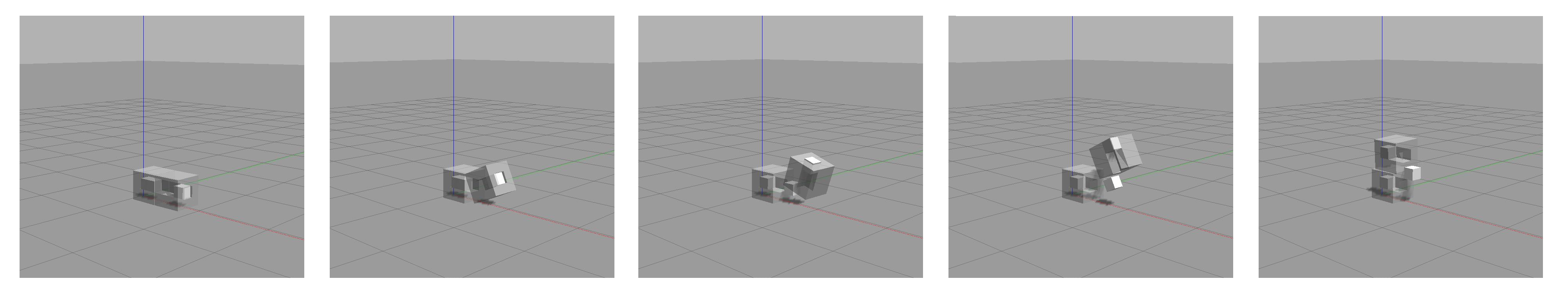}
    \caption{Lifting}
    \label{fig:lift}
\end{figure}

Lifting is, similarly to turning, a combination of translational and rotational motion, see Fig.~\ref{fig:lift}.

\section{Planning}
\label{sec:planning}
Planning in MoleMOD is based on the A* algorithm -- an informed method for state-space search.
A state is represented by an arrangement of passive blocks and robots positions.
We assuming two-dimensional space, which can be described as a two-dimensional matrix, where each cell stores the information whether the corresponding space is empty, it contains a block, or a block with a robot. 

An action is performed by movement of robots, which can relocate some blocks. 
The simplest action is moving from one block to another. The natural condition for this action is that the position, where the robot moves to, is neighboring to the current robot position, lies inside a given area and also it contains an empty block. The robot can move to the right, to the left, up and down this way.

Another type of movement is lifting of a block. In reality, the model expands and partially inserts into the blocks beside. After that, the robot lifts one block up and place it on the top of the second one. Finally, the model retracts into one of the two blocks. To reduce the number of possibilities, we assume that the robot picks up the block, where it originally was, and remains in the block after the movement. To the motion can be executed, there has to be a free space around the moving block.

Putting down is similar to lifting. The robot expands into the block under its position and then contracts with the top one. For simplicity, the model starts and finishes again in the moved block.

The last motion that is possible with only one robot is moving a block to the right or the left. The model is in the block that we want to shift. If the place next to the block is free and under it, there is another block, the robot expands to this block. Then it moves the block to the empty position and contracts. The moving finishes again in the shifted block.

All movements, which were mentioned above, are valid also for the case with more robots. The advantage is that they can be performed in parallel, so the entire construction is done faster.

\begin{figure}[htbp]
	\centering
	\includegraphics[width=\textwidth]{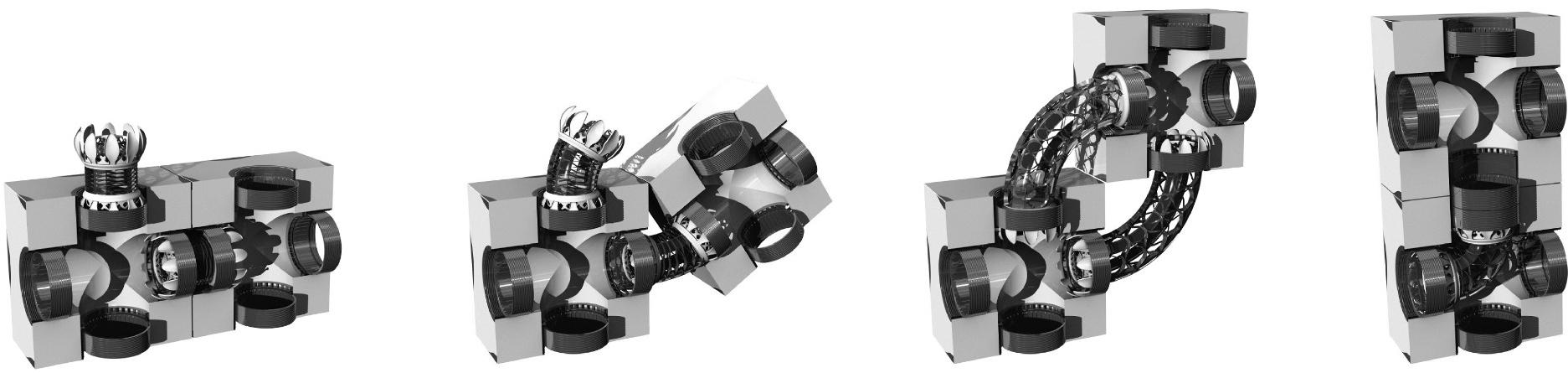}
	\caption{Cooperation of two robots}
	\label{fig:two}
\end{figure}

Besides, robots can work on a movement together. A block is lifted by one robot to a certain position, where a second robot takes it and completes the move as shown in Fig.\ref{fig:two}.

The planning algorithm applies one action, where cooperation of two robots is used. When the block is lifted by one robot, it is like stairs, the block moves not only up but also to a side. In some cases, however, it is necessary to move just upwards.
The movement starts identically as lifting. The robot in the block expands to the block beside and then elevates the block one position up. The second robot has to be in the block that lies on the block where the robot stretched to before. The second robot expands and ``catches'' the block and the first robot contracts to the underlying block.

\subsection{Cost calculation}
The evaluation function of A* has a standard form $f(x) = g(x) + h(x)$, where $g(x)$ is the cost of the path from the start node to $x$, and $h(x)$ is a heuristic function that estimates the cost of the cheapest path from $x$ to the goal. 

One possibility of computing a value of the cost $g$ is based on distance of the start and current states: the distance of a robot or a block moving is equal to $|i'- i| + |j' - j|$, where $(i, j)$ are coordinates of the robot or the robot in the first state and $(i', j')$ are its coordinates in the second state. However, computing the distance in every step means to find the robot or block that has just been moved, and calculate how far has shifted. That is senselessly complicated. Because the number of movements is limited and each has a specific distance that never changes, it is much simpler to assign a value $d$ to every movement $m$:

$$g(x_{n}) = g(x_{n-1}) + d(m).$$

While the case of one robot is simple as the robot can perform only one movement in one step,  the case with more robots is more complicated. For example, two robots can shift two blocks at the same time or gradually in two steps. The second option does not have any advantages, it only prolongs the building, so it is necessary to obviate it. That can be done by adding 1 to the distance in every step:

$$g(x_{n}) = g(x_{n-1}) + d(m) + 1,$$

where $d(m)$ is the sum of costs of the motions that have been made to get from the state $x_{n-1}$ to the state $x_{n}$.

Counting the distance to the goal state is more difficult. A robot moves to a block, shifts it, moves to another block, shifts it and so over and over again until the goal state is reached. It is almost impossible to calculate the real cost.

When a block moves between two states, the cost is equal to $|i'- i| + |j' - j|$. If more blocks are moved, the cost is $\sum_{n=1}^{N}|i'_{n}- i_{n}| + |j'_{n} - j_{n}|$, where $N$ is the number of blocks that have been moved, and $[i_{n}, j_{n}]$ and $[i'_{n}, j'_{n}]$ are coordinates of $n$-th block before and after the movement.

Costs of robots movement have to be added to the total cost, but it is hard to reckon them in advance. It depends on the number of robots and blocks and their positions. Every time a robot moves a block to a correct location,it has to move to the position next to the block. The minimal cost of the path from a block to another is one and the total cost of the way between blocks is equal to $n_{wrong}$, where $n_{wrong}$ is the number of the blocks in the wrong positions which do not contain an robot. Therefore, the estimation of the cost of the path from the current state $x$ to the goal state $x_{G}$ is:

$$h(x) = \sum_{n=1}^{n_{wrong}}(|i_{Gn} - i_{n}| + |j_{Gn} - j_{n}|) + n_{wrong} - n_{robot},$$

where $[i_{Gn},j_{Gn}]$ is a goal position of $n$-th block.

\subsection{Experiments}

\begin{figure}[htbp]
	\centering
	\includegraphics[width=\textwidth]{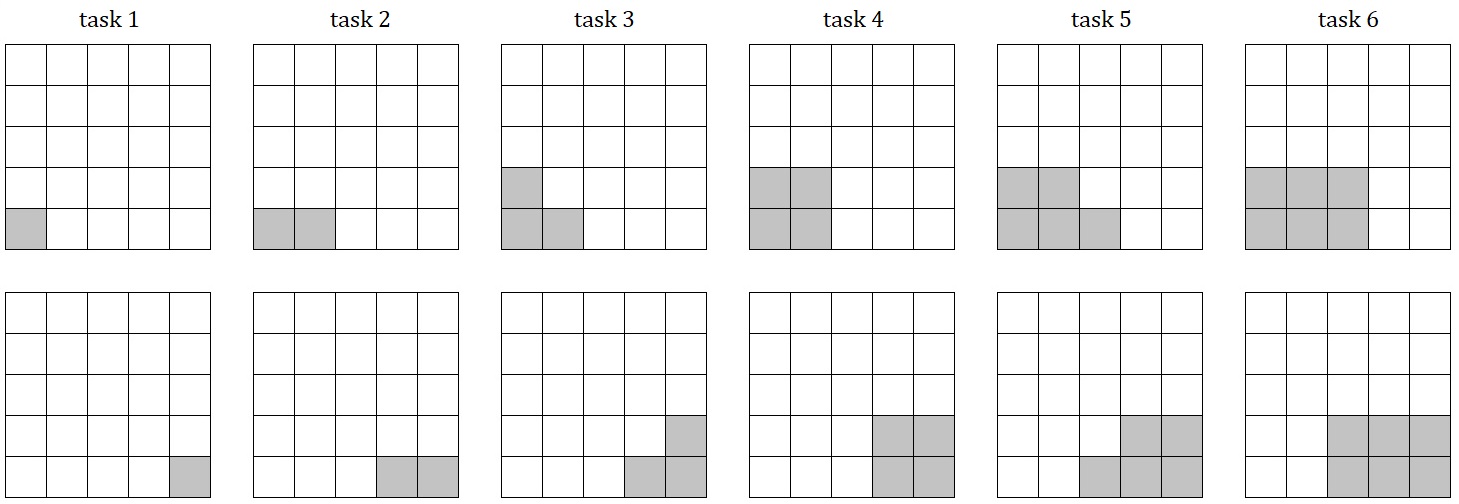}
	\caption{Planning tasks for time measurement}
	\label{fig:dg1}
\end{figure}

Several experiments have been performed to demonstrate feasibility and time complexity of the planning algorithm.
Six planning tasks depicted in Fig.~\ref{fig:dg1} were run. In each pair, the upper state represents the initial layout of the blocks and the state below the goal arrangement. All tasks have been successively solved for different numbers of robots and for all cases the time of the planning was measured. 

\begin{table}[htbp]
	\centering
	\caption{Time of planning}
	\begin{tabular}{|c|c|c|c|c|c|c|}
		\hline
		\multicolumn{7}{|c|}{time [ms]}\\ \hline
		& 1 robot & 2 robots & 3 robots & 4 robots & 5 robots & 6 robots\\ \hline
		task 1 & 0.24 & & & & &\\ \hline
		task 2 & 1.87 & 2.02 & & & &\\ \hline
		task 3 & 3.08 & 18.23 & 5.10 & & &\\ \hline
		task 4 & 9.00 & 75.60 & 117.95 & 9.91 & &\\ \hline
		task 5 & 15.50 & 224.10 & 614.93 & 332.15 & 18.63 &\\ \hline
		task 6 & 35.64 & 1212.75 & 4390.14 & 4000.69 & 706.35 & 22.48\\ \hline
    \end{tabular}
    
	\label{tbl:dg1}
\end{table}

Table~\ref{tbl:dg1} shows that the time needed to find a solution, if the system contains more robots, is significantly higher than the case of one robot. However, the ratio of blocks and robots is also important. Robots need sufficient amount of space to move, so in the case that most blocks are occupied, the number of options is reduced. The other case is, if the quantity of blocks is quite higher than the number of robots, so the robots have plenty of space for moving. Each robot can perform some movements, the number of moves depends on the specific conditions. One more robot adds its motions and combinations of its moves and moves of others. The larger space and the more blocks it contains, the more movements will be possible.

\section{Conclusion}
\label{sec:conclusion}

In this paper we presented MoleMOD -- heterogeneous self-reconfigurable modular robotic system and two software parts of it -- the simulation environment and the planning algorithm. 
The system is still in the first phase of development and thus only first results were presented which can be further improved.
Especially the planning algorithm has a potential for next improvements. 
First, better heuristics, which more precisely estimates the cost to the goal, needs to be designed.
This will guide the A* algorithm to search the state space more effectively by expanding less states. 
Secondly, we will investigate more advanced planning algorithms (e.g., hierarchical) which will further improve computational complexity of the planning.
Finally, extension of the algorithm into the 3D case will be done.
Regarding the simulation environment, we would like to equip robots with sensors to accelerate development of control algorithms for robots. 

\section*{Acknowledgement}
This work has been supported by the European Union's Horizon 2020 research and innovation programme under grant agreement No 688117, and by the European Regional Development Fund under the project Robotics for Industry 4.0 (reg. no. CZ.02.1.01/0.0/0.0/15 003/0000470).
        
\bibliographystyle{splncs04}
\bibliography{main}
    
\end{document}